\documentclass[runningheads]{llncs}
\usepackage{graphicx}
\usepackage{times}
\usepackage{mathptmx}
\usepackage{multirow}
\usepackage{float}

\usepackage{graphicx}
\usepackage{fancyhdr}
\fancypagestyle{specialfooter}{%
  \fancyhf{}
  
  \fancyhead[R]{\scriptsize{Accepted in CVMI 2022 Conference}}
  \fancyfoot[L]{\scriptsize{International Conference on Computer Vision and Machine Intelligence (CVMI), 2022}}
}

\begin{document}

\title{Context Unaware Knowledge Distillation for Image Retrieval}


\author{Bytasandram Yaswanth Reddy\inst{1} \and
Shiv Ram Dubey\inst{2} \and Rakesh Kumar Sanodiya\inst{1} \and
Ravi Ranjan Prasad Karn\inst{1}}
\authorrunning{B. Yaswanth et al.}
\institute{Department of Computer Science and Engineering, Indian Institute of Information Technology, Sri City, Chittoor, India \and
Computer Vision and Biometrics Laboratory, Indian Institute of Information Technology, Allahabad, India\\
\email{yaswanthreddy.b18@iiits.in, srdubey@iiita.ac.in, rakesh.s@iiits.in, raviranjanprasad.k@iiits.in}}

\maketitle
\thispagestyle{specialfooter}

\begin{abstract}
Existing data-dependent hashing methods use large backbone networks with millions of parameters and are computationally complex. Existing knowledge distillation methods use logits and other features of the deep (teacher) model and as knowledge for the compact (student) model, which requires the teacher's network to be fine-tuned on the context in parallel with the student model on the context. Training teacher on the target context requires more time and computational resources. In this paper, we propose context unaware knowledge distillation that uses the knowledge of the teacher model without fine-tuning it on the target context. We also propose a new efficient student model architecture for knowledge distillation. The proposed approach follows a two-step process. The first step involves pre-training the student model with the help of context unaware knowledge distillation from the teacher model. The second step involves fine-tuning the student model on the context of image retrieval. In order to show the efficacy of the proposed approach, we compare the retrieval results, no. of parameters and no. of operations of the student models with the teacher models under different retrieval frameworks, including deep cauchy hashing (DCH) and central similarity quantization (CSQ). The experimental results confirm that the proposed approach provides a promising trade-off between the retrieval results and efficiency. The code used in this paper is released publicly at \url{https://github.com/satoru2001/CUKDFIR}.

\keywords{Knowledge Distillation \and Image Retrieval \and CNN Model \and Model Compression.}
\end{abstract}
\section{INTRODUCTION}
In this age of Big data, where voluminous data is generated from various sources with very fast speed, for image retrieval-based applications, parallel to the indexing methods \cite{lew2006content}, hashing methods have shown promising results. In the hashing methods, high dimension media data like images/video are compressed into a low dimension binary code (hash) such that media with similar data items have identical hash. Deep learning has become successful in various fields in the past few years. Many state-of-the-art models \cite{VTS,fdresnet} emerged with a large number of trainable parameters, making them good at learning complex patterns from data and computationally costly to run on low-end devices. Many model compression and acceleration techniques were introduced to decrease the computational complexity, like parameter tuning/quantization, transferred/compact convolution filters, Low-rank factorization, and Knowledge Distillation \cite{cheng2017survey}. Knowledge distillation is a model compression and acceleration method that helps the compact (student) model to perform nearly equal to or better than the deep (teacher) models by learning knowledge gained by deep (teacher) models. In vanilla Knowledge distillation \cite{hinton2015distilling,DBLP:journals/corr/BaC13} logits of the pre-trained teacher model on the context are used as knowledge for the student network. 

Knowledge distillation can be categorized into three types \cite{DBLP:journals/corr/abs-2006-05525}. First, Response based knowledge distillation in which the logits of teachers are used as knowledge for students. Hilton et al. \cite{hinton2015distilling} used soft targets, which are soft-max probabilities of classes in which input is predicted to be ascertained by the teacher model as knowledge to distill. Second, Feature-based knowledge distillation in which both the features of the last and intermediate layers are used as knowledge \cite{romero2014fitnets}. Third, Relation-based knowledge, instead of learning from the features of intermediate layers/logits like in previous models, relation-based knowledge distillation tries to distill the relationship between different layers or data samples. Yim et al. \cite{yim2017gift} used a Gram matrix between two layers that are calculated using the inner products between features from both the layers. It summarizes the relationship between the two layers and tries to distill this knowledge into the student. There are two main schemas of training during knowledge distillation, online and offline schemas. In the offline schema, we have fine-tuned teacher networks, and the student network will learn the teacher's knowledge along with the downstream task. In an online schema, the teacher and the student are trained simultaneously on the downstream task and knowledge distillation. A great deal of comprehensive survey is conducted on this in \cite{DBLP:journals/corr/abs-2006-05525}.   

It has been seen that, recently, Deep learning to hash methods enables end-to-end representation learning. They can learn complex non-linear hash functions and achieve state-of-the-art retrieval performance. In particular, they proved that the networks could learn the representations that are similarity-preserved and can quantize representations to be binary codes  \cite{cao2018deep,yuan2020central,zhu2016deep,Cao2016DeepQN,liu2018deep,zhai2020deep,cao2017hashnet}. Deep learning based hashing can be categorized into various buckets based on the training approach, namely supervised, unsupervised, semi-supervised, weakly supervised, pseudo supervised, etc. The backbone architectures used for these different training modes contain CNN, Auto-encoders, Siamese, Triplet networks, GAN, etc., and use different descriptors for representing hash code, namely Binary, where hashes are a combination of 0's and 1's, real-valued, and aggregation of both binary and real-valued. A comprehensive survey on image retrieval is conducted in \cite{dubey2021decade} which can be referred for holistic understanding. 

Although most existing deep learning methods for hashing are tailored to learn efficient hash codes, the used backbone models become computationally costly with millions of parameters. On the other hand, most existing knowledge distillation modes are carried out in two steps; the first step consists of fine-tuning the teacher model on the context, followed by the training of the student model with knowledge distillation and contextual losses. However, constantly fine-tuning teacher models for each context might be computationally expensive since they tend to be deeper networks. 

In this work, we present the findings on the approach for context unaware knowledge distillation in which the knowledge is transferred from a teacher network that is not fine-tuned on the context/downstream task. This approach is carried out in two steps; first, we distill the knowledge from an un-trained teacher (on context) to a student on a specific dataset making the student mimic the output of the teacher for that dataset. Then we fine-tune our student on any context of the same dataset, thus decreasing the computation overhead incurred from fine-tuning the teacher network for each context on the dataset. We experimented our approch in the context of image retrieval, then compared the results of teacher and student by using each of them as a backbone network in CSQ \cite{yuan2020central} and DCH \cite{cao2018deep}. The experiments are conducted on the Image retrieval task on two different datasets with multiple hash lengths. 

\section{\uppercase{Related Work}}
The supervised deep learning to hash methods such as DCH \cite{cao2018deep}, CSQ \cite{yuan2020central}, DTQ \cite{liu2018deep}, DQN \cite{Cao2016DeepQN}, DHN \cite{zhu2016deep} are successful in learning non-linear hash function for generating hash of different bit-sizes (16, 32, 48, 64...) and achieved state-of-the-art results in image retrieval. DQN \cite{Cao2016DeepQN} uses a Siamese network that uses pairwise cosine loss for better linking the cosine distance of similar images and a quantization loss for restricting the bits to binary. DTQ \cite{liu2018deep} introduces the Group Hard triplet selection module for suitable mining triplets (anchor, positive and negative) in real-time. The concept behind it is to divide the training data into many groups at random, then pick one hard negative sample for each anchor-positive pair from each group at random. A specified triplet loss is used for pulling together anchor and positive pairs and moving away anchor and negative pairs, as well as a quantization loss for monitoring the efficiency and restricting hash bits to be binary.

DCH \cite{cao2018deep} exploits the lack of the capability of concentrating relevant images to be within a small Hamming distance of existing hashing methods. Instead, it uses Cauchy distribution for distance calculation between similar and dissimilar image pairs instead of traditional sigmoid. Similar to previous methods, DCH \cite{cao2018deep} includes quantization loss for controlling hash quality. CSQ \cite{yuan2020central} replaces the low efficiency in creating image datasets while using pairwise or triplet loss by introducing a new concept of ``hash centers". Hash centers are unique $K$ dimension vectors ($K$ refers to desired hash size), one for each class. They are defined as $K$-dimensional binary points (0/1) in hamming space with an average pairwise distance greater than or equal to $K/2$ between any two hash centers. Now they use these hash centers as a target, similar to a multi-class classification problem.

DTQ \cite{zhai2020deep} uses a compact student network for fast image retrieval. The training is carried out in three phases. In the first phase, a modified teacher is trained on the classification task. The teacher and student models have a fully connected layer of $N$ neurons ($FC@N$) (where $N$ indicates desired hash length) before the classification layer. The second phase consists of knowledge distillation between teacher and student, where the output of the teacher's $FC@N$ is used as knowledge. The loss function for the student network includes knowledge distillation loss (a regression loss between the teacher and student's output of $FC@N$) and the classification loss. In the last phase, the full precision student model is quantized to get a ternary model (where weights of each layer are represented in only three states) and fine-tuned with knowledge distillation to find the best ternary model.

\begin{figure}[!t]
  \centering
    \includegraphics[width=\textwidth]{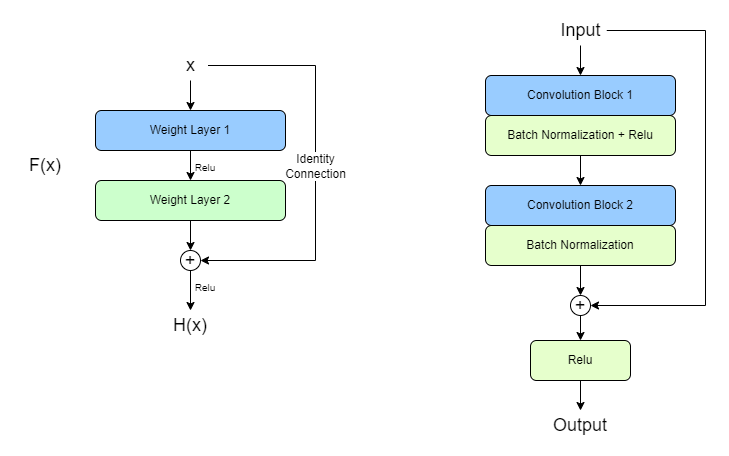}
    \caption{(left) Residual block of ResNet architecture \cite{he2016deep} and (right) Building block of Student architecture.}
    \label{fig:basicblock}
 \end{figure}

\section{\uppercase{Context Unaware Knowledge Distillation}}
In this section, we first present an overview of ResNet followed by the architecture of two student models V1 and V2, one for each teacher, namely ResNet-50 \cite{he2016deep} and AlexNet \cite{krizhevsky2012imagenet}, respectively. We then present the process of knowledge distillation between teacher and student.

\subsection{ResNet Overview}
The building block of ResNet is the residual block which is one of the significant advancements in deep learning. The problem with general plain deep learning architectures, which constitute a sequence of convolution layers and other layers like batch normalization, etc., is diminishing gradient during backpropagation. Weights of specific layers cannot be updated since no gradient affects the model's learning and degrades its performance. Residual blocks tackle this problem by adding an identity connection which acts as a way to avert the vanishing gradient problem. Let us denote $H(x)$ as the desired mapping function where $x$ is the input of the residual block. We make the residual building block fit the mapping function $F(x)$ such that $F(x) = H(x) - x$. It is demonstrated in the left subfigure of Fig~\ref{fig:basicblock}.

\subsection{Student Model}
Each student model consists of two building blocks, namely Basic Block, which is inspired by Residual blocks \cite{he2016deep}. These blocks are stacked to form a layer. These layers are then stacked together to form the student network.

\subsubsection{Basic Block and Layer}
\begin{figure*}[!t]
  \centering
    \includegraphics[width=\textwidth]{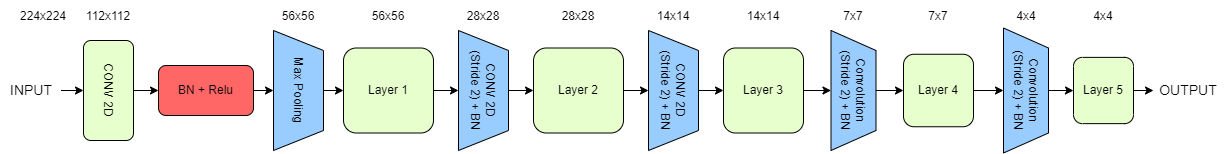}
    \caption{Represents Student Model, \textbf{$X\times{X}$} on top of each blocks represents the dimension of output from that layer (excluding channel dimension). Each layer is followed by a convolution layer (CONV2D) with stride of two to decrease the dimensions by half instead of using max-pooling. BN stands for Batch Normalization.}
    \label{fig:model}
 \end{figure*}

The architecture of the basic block is demonstrated in the right subfigure of Fig~\ref{fig:basicblock}. Each convolution layer contains a kernel of size $3\times3$ and a stride of one. The input dimension is retained throughout the basic block to support the identity connection.    
Each student model contains five layers, and each of the five layers contains $2$, $3$, $5$, $3$, and $2$ basic blocks stacked together, respectively. The dimension of the input feature is retained in a layer. Each layer (except the $5^{th}$) is followed by a convolution block with a $3\times3$ kernel and stride of two to reduce the dimension of the features by half, then followed by a batch normalization layer. The architecture of the student is demonstrated in Fig~\ref{fig:model}. Each layer and the following convolution block contain an equal number of filters.

\subsubsection{Student Network}
Since the dimensions of flattened output features before classification layer of ResNet-50 \cite{he2016deep} is 2048 and that of Alexnet \cite{krizhevsky2012imagenet} is 4096, we made two different student models \textbf{StudentV1} and \textbf{StudentV2}. Before passing input to the first layer, we tried to reduce the dimension of the input similar to ResNet \cite{he2016deep}. The initial module of student consists of a convolution block with a $7\times7$ filter, stride of two, and padding of three. It is then followed by Batch normalization and ReLU to reduce the dimensionality of input by half, followed by a max-pooling layer of stride two to reduce the dimension of features by half further. The architectures of the two student models differ only in the number of filters in $5^{th}$ layer.   
\textbf{StudentV1} is the student model for ResNet-50 \cite{he2016deep}. The layer wise summary of StudentV1 is demonstrated in Table~\ref{tab:student v1}. \textbf{StudentV2} is the student model for Alexnet \cite{krizhevsky2012imagenet}. The layer wise summary of StudentV2 is demonstrated in Table~\ref{tab:student v2}     

\subsubsection{Comparison}
The comparison of teacher and their respective student models are done in Table~\ref{tab:comparision1} and Table~\ref{tab:comparision2}, respectively. With less trainable parameters and fewer FLOPs, the model takes less time to train per epoch as well as for inference. We can observe that $85.46\%$ reduction in the number of trainable parameters in the ResNet50-StudentV1 pair and a $91.16\%$ reduction in the number of trainable parameters in the AlexNet-StudentV2 pair. In the AlexNet-StudentV2 pair, as StudentV2 contains identity connection whereas AlexNet doesn't, it results in higher FLOPs for Student despite having lower trainable parameters. In the ResNet50-StudentV1 pair, since both networks contain identity connections, fewer trainable parameters in student resulted in lower FLOPs.

\begin{table}[!t]
\caption{The layerwise summary of StudentV1 with input size of $224\times224\times3$.}
\label{tab:student v1} 
\centering
\begin{tabular}{c c c}
  \hline
  Layer & Number of filters & Output Shape  \\
  \hline
  InputLayer & $0$ & $224\times224\times3$ \\
  Initial Module (Conv 2D + BN + ReLU) & $64$ &  $112\times112\times64$\\
  Max Pool & $64$ &  $56\times56\times64$\\
  Layer\_1 & $64$ &  $56\times56\times64$\\
  Conv\_2d\_1 (Conv 2D + BN) & $64$ &  $28\times28\times64$\\
  Layer\_2 & $64$ &  $28\times28\times64$\\
  Conv\_2d\_2 (Conv 2D + BN) & $128$ &  $14\times14\times128$\\
  Layer\_3 & $128$ &  $14\times14\times128$\\
  Conv\_2d\_3 (Conv 2D + BN) & $128$ &  $7\times7\times128$\\
  Layer\_4 & $128$ &  $7\times7\times128$\\
  Conv\_2d\_4 (Conv 2D + BN) & $128$ &  $4\times4\times128$\\
  Layer\_5 & $128$ &  $4\times4\times128$\\
  Flatten & NA & $2048$ \\
  \hline
  *BN stands for Batch Normalisation
\end{tabular}
\end{table}

\begin{table}[!t]
\caption{The layerwise summary of StudentV2 with input size of $224\times224\times3$.}
\label{tab:student v2}
\centering
\begin{tabular}{c c c}
  \hline
  Layer & Number of filters & Output Shape  \\
  \hline
  InputLayer & $0$ & $224\times224\times3$ \\
  Initial Module (Conv 2D + BN + ReLU) & $64$ &  $112\times112\times64$\\
  Max Pool & $64$ &  $56\times56\times64$\\
  Layer\_1 & $64$ &  $56\times56\times64$\\
  Conv\_2d\_1 (Conv 2D + BN) & $64$ &  $28\times28\times64$\\
  Layer\_2 & $64$ &  $28\times28\times64$\\
  Conv\_2d\_2 (Conv 2D + BN) & $128$ &  $14\times14\times128$\\
  Layer\_3 & $128$ &  $14\times14\times128$\\
  Conv\_2d\_3 (Conv 2D + BN) & $128$ &  $7\times7\times128$\\
  Layer\_4 & $128$ &  $7\times7\times128$\\
  Conv\_2d\_4 (Conv 2D + BN) & $128$ &  $4\times4\times128$\\
  Layer\_5 & $256$ &  $4\times4\times256$\\
  Flatten & NA & $4096$ \\
  \hline
  *BN stands for Batch Normalisation
\end{tabular}
\end{table}

\begin{table}[!t]
\caption{Comparison between ResNet-50 and StudentV1. FLOPs are calculated on image of dimension $224\times224\times3$}
\label{tab:comparision1} 
\centering
\begin{tabular}{|c|c|c|}
  \hline
  Model & Trainable Parameters & FLOPs  \\
  \hline
  Resnet50 & $23,639,168$ & $4.12$ Giga \\
  \hline
  StudentV1 & $3,437,568$ & $1.10$ Giga \\
  \hline
\end{tabular}
\end{table}

\begin{table}
\caption{Comparison between AlexNet and StudentV2. FLOPs are calculated on image of dimension $224\times224\times3$.}
\label{tab:comparision2} 
\centering
\begin{tabular}{|c|c|c|}
  \hline
  Model & Trainable Parameters & FLOPs  \\
  \hline
  Alexnet & $57,266,048$ & $0.72$ Giga \\
  \hline
  StudentV2 & $5,060,352$ & $1.119$ Giga \\
  \hline
  \end{tabular}
\end{table}
 
\subsection{Knowledge Distillation}
\begin{figure}[!t]
  \centering
    \includegraphics[width=4.5cm, height=4cm]{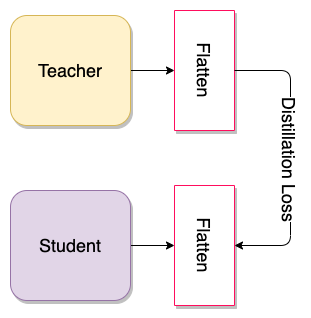}
    \caption{Training process for knowledge distillation in which we freeze the weights of teacher model and train student model with MSE based knowledge distillation loss.}
    \label{fig:kdlos}
 \end{figure}

We can view the process of knowledge distillation from teacher to student network as a regression problem. Hence, we can use the loss function of regression problems like L1 (mean absolute error, mae), L2 (mean squared error, mse), or smooth L1 loss. We consider Mean Square Error (MSE) as our loss function as most of the activation value of the last layer is less than 1, which makes smooth L1 loss perform similar to L2 loss. Here we use the output features of the last layer of the teacher as knowledge to train the student. The equation is given as follows,
\begin{equation}\label{eq1}
   L_{2}\left (T,S \right) = \frac{1}{N}\sum_{i=1}^{N}\sum_{j=1}^{K}(S_{ij}-T_{ij})^{2}
\end{equation}
where $N$ represents the number of images in the dataset, $K$ represents the dimensionality of features which is 2048 and 4096 for ResNet-50 \cite{he2016deep} and AlexNet \cite{krizhevsky2012imagenet}, respectively, $T_{ij}$ represents the feature vectors of the last layer of the Teacher network, and $S_{ij}$ represents the feature vector of the last layer of the student network. We fix the teacher weights and only update the student weights during back-propagation as portrayed in Fig. \ref{fig:kdlos}.

\begin{table*}[t]
\caption{Comparison of mAP of different bits under CSQ retrieval framework \cite{yuan2020central}.} \label{tab:metric1} 
\centering
\begin{tabular}{|l|l|l|l|l|l|l|}
\hline
\multirow{2}{*}{Model} & \multicolumn{3}{l|}{NUS-WIDE(mAP@5000)} & \multicolumn{3}{l|}{CIFAR10(mAP@5000)} \\ \cline{2-7} 
          & 16 bit & 32 bit & 64 bit & 16 bit & 32 bit & 64 bit \\ \hline
Resnet50  & 0.812  & 0.833  & 0.839  & 0.834  & 0.851  & 0.849  \\ \hline
StudentV1 & 0.779  & 0.807  & 0.819  & 0.824  & 0.822  & 0.840  \\ \hline
Alexnet   & 0.762  & 0.794  & 0.808  & 0.784  & 0.778  & 0.787  \\ \hline
StudentV2 & \textbf{0.765}  & \textbf{0.798}  & \textbf{0.812}  & 0.763  & 0.747  & 0.767  \\ \hline
\end{tabular}
\end{table*}

\begin{table*}[t]
\caption{Comparison of mAP of different bits under DCH retrieval framework \cite{cao2018deep}}
\label{tab:metric2} 
\centering
\begin{tabular}{|l|l|l|l|l|l|l|}
\hline
\multirow{2}{*}{Model} & \multicolumn{3}{l|}{NUS-WIDE(mAP@5000)} & \multicolumn{3}{l|}{CIFAR10(mAP@5000)} \\ \cline{2-7} 
          & 16 bit & 32 bit & 48 bit         & 16 bit & 32 bit & 48 bit \\ \hline
Resnet50  & 0.778  & 0.784  & 0.780          & 0.844  & 0.868  & 0.851  \\ \hline
StudentV1 & 0.766  & 0.781  & \textbf{0.782} & 0.819  & 0.828  & 0.845  \\ \hline

Alexnet   & 0.748  & 0.76   & 0.758          & 0.757  & 0.786  & 0.768  \\ \hline

StudentV2 & 0.743  & 0.755  & \textbf{0.762} & 0.754  & 0.769  & 0.754  \\ \hline
\end{tabular}
\end{table*}

\begin{figure}[t]
  \centering
    \includegraphics[width=\textwidth]{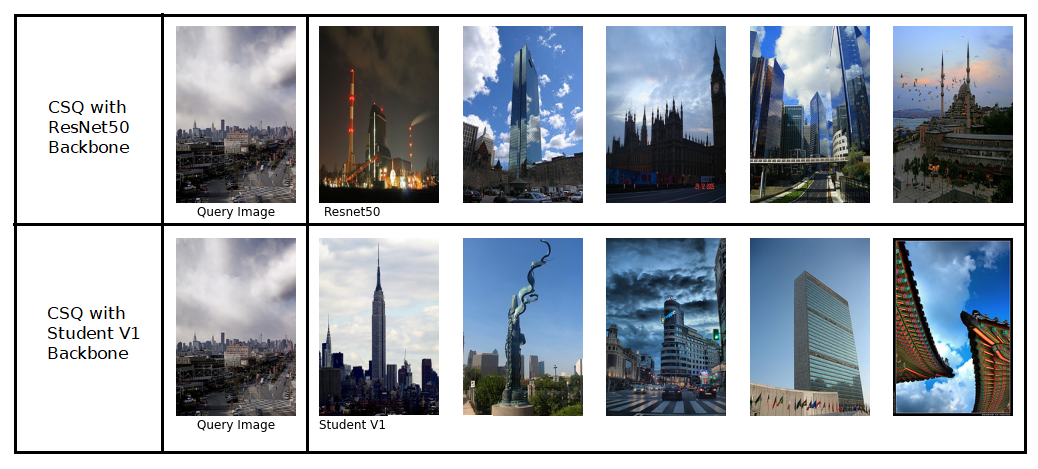}
    \caption{The top row shows the retrieved images using ResNet50 \cite{he2016deep} as the backbone in CSQ \cite{yuan2020central} with a 64-bit hash, and the bottom row shows the retrieved images using StudentV1 as a backbone in CSQ with a 64-bit hash. The left subfigure in each row represents the Query Image from the test set of NUS-WIDE \cite{chua2009nus}, and the following subfigures represent the top 5 similar images from the Database.}
    \label{fig:res_ret}
 \end{figure}
 
 \begin{figure}[t]
  \centering
    \includegraphics[width=\textwidth]{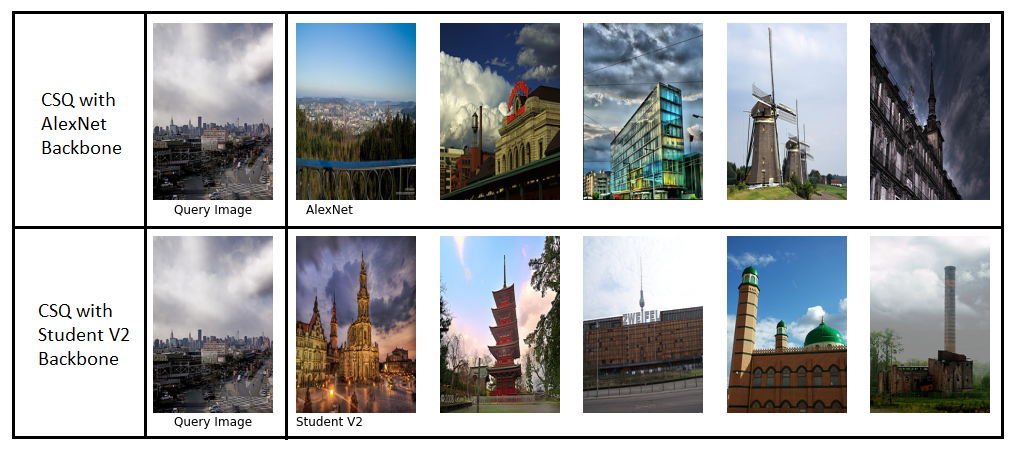}
    \caption{The top row shows the retrieved images using AlexNet \cite{krizhevsky2012imagenet} as the backbone in CSQ \cite{yuan2020central} with a 64-bit hash, and the bottom row shows the retrieved images using StudentV2 as a backbone in CSQ with a 64-bit hash. The left subfigure in each row represents the Query Image from the test set of NUS-WIDE \cite{chua2009nus}, and the following subfigures represent the top 5 similar images from the Database.}
    \label{fig:alex_ret}
 \end{figure}
 
\begin{table}[!t]
    \caption{Comparison between the training time per epoch (in seconds) of students and their respective teachers on NUS-WIDE dataset}
    \label{tab:comparision3} 
    \centering
    \begin{tabular}{|c|c|c|c|c|}
      \hline
      Loss & ResNet50 & StudentV1 & AlexNet & StudentV2  \\
      \hline
      CSQ 64 & 118 & 37 & 40 & 38\\
      \hline
      DCH 48 & 120 & 39 & 39 & 38\\
      \hline
    \end{tabular}
    \par
    {All the experiments are conducted on Tesla T4 GPU and Intel Xeon CPU. \par}
\end{table}

\section{\uppercase{EXPERIMENTS}}
As discussed above, the first step includes training the teacher-student pair with knowledge distillation loss. We then fine-tune the student model using the pre-trained student model as the backbone network instead of their respective teacher on image retrieval task under the retrieval frameworks of CSQ \cite{yuan2020central} and DCH \cite{cao2018deep}.

\subsection{Datasets and Evaluation Metrics}
We use the CIFAR10 \cite{krizhevsky2009learning} and NUS-WIDE \cite{chua2009nus} datasets for experiments. CIFAR10 \cite{krizhevsky2009learning} contains images from $10$ different classes (categories) and each class include $6,000$ images. For knowledge distillation, we use the entire dataset for training teacher-student pairs. For fine-tuning on image retrieval task, we randomly select $1,000$ images ($100$ images per class) as the query set and $5,000$ images as the training set ($500$ images per class), with the remaining images as database images as done in the works of DCH \cite{cao2018deep} and CSQ \cite{yuan2020central}.
NUS-WIDE \cite{chua2009nus} is a public Web image dataset that contains $2,69,648$ images. We use the subset of NUS-WIDE in which there are only $21$ frequent categories. We use the entire NUS-WIDE dataset to train teacher-student pairs for knowledge distillation. For fine-tuning on the image retrieval task, we randomly choose $2,100$ images ($100$ images per class) as a test set and $10,500$ images ($500$ images per class) as a training set, leaving the rest $1,49,736$ as a database. 

We use mean average precision ($mAP$) as the evaluation metric for image retrieval. To calculate $mAP@N$ for a given set of queries, we first calculate Average Precision $(AP)@N$ for each set of the query as specified as follows,
\begin{equation}\label{eq2}
  AP@N=  \frac{\sum_{i=1}^{N} P(i) \alpha(i)}{\sum_{i=1}^{N} \alpha(i)}
\end{equation}
where $P(i)$ is precision of $i^{th}$ retrieved image and $\alpha(i) = 1$ if the retrieved image is a neighbor (belongs to same class) and $\alpha(i) = 0$ otherwise. $N$ denotes number of images in the database.
$mAP$ is calculated as mean of each query average precision and is represented as follows,
\begin{equation}\label{eq3}
  mAP@N=  \frac{\sum_{i=1}^{Q} AP@N(i)}{Q}
\end{equation}
where $Q$ represents a number of query images. We use $mAP@5000$ as our evaluation metric for the image retrieval for both datasets.    

\subsection{Training and Results}
Both the teacher networks are initialised with the pre-trained weights of ImageNet \cite{imagenet} classification task. During knowledge distillation training, the Adam optimizer is used with learning rate (LR) of 1e-4 for the ResNet50-StudentV1 pair and LR of 3e-6 for the training of the AlexNet-StudentV2 pair. We train teacher-student pairs for 160 and 120 epochs for CIFAR10 \cite{krizhevsky2009learning} and NUS-WIDE \cite{chua2009nus} datasets, respectively. We add a Fully Connected layer with $n$ neurons (where $n$ is the number of desired hash bits such as $16$,$32$,$48$,$64$) to the student and teacher models for image retrieval. We then train it under the retrieval frameworks of CSQ \cite{yuan2020central} and DCH \cite{cao2018deep}. For training on image retrieval, we use the RMSProp optimizer with a learning rate of 1e-5. Table~\ref{tab:metric1} and Table~\ref{tab:metric2} represent the results with backbone networks as teachers (i.e., Resnet50 and Alexnet), students (i.e., StudentV1 and StudentV2) for different hash bits under CSQ and DCH retrieval frameworks, respectively.
It can be seen that the performance of StudentV1 and StudentV1 models are either better or very close to Resnet50 and Alexnet teacher models, respectively, in spite of having significantly reduced number of parameters. Moreover, the no. of FLOPS of StudentV1 model is also significantly as compared to the Resnet50 model. It is also noted that the StudentV2 model outperforms the Alexnet model on NUS-WIDE dataset under CSQ framework. The performance of the proposed student models are better for 48 bit hash codes on NUS-WIDE dataset under DCH framework. Table~\ref{tab:comparision3} represents the training time per epoch of teacher with their respective students (in seconds) on NUS-WIDE dataset for CSQ 64 bit configuration and DCH 48 bit configuration, we can observe a drop of nearly $3.1X$ and $1.05X$ times training time per epoch in the case of StudentV1 and StudentV2 respectively.

For a given Query image from the test set, Fig~\ref{fig:res_ret} and Fig~\ref{fig:alex_ret} represent the top five retrieved images from the database of $1,49,736$ images based on Hamming distance. The Query image belongs to three categories: Building, Cloud, and Sky. All the retrieved images contain at least two categories, i.e., Building and Sky making them relevant retrieved images.

\section{\uppercase{Conclusions}}
\label{sec:conclusion}
Deep learning to hash is an active research area for image retrieval tasks where one uses deep learning algorithms to act as hash functions with appropriate loss functions such that images with similar content has a similar hash. Most present-day deep learning algorithms have deep convolution neural networks as backbone models, which are computationally expensive. In general, compact (student) models with less trainable parameters are less computationally complex, but do not perform well as deep (teacher) models on the tasks. Most existing knowledge distillation methods require the teacher model to be fine-tuned on the task, requiring more training time and computational resources. In this work, we propose a two-fold solution to increase the performance of the student model using knowledge from a teacher which is not trained on the context.

We observe that the student model performed in equal terms to teacher models with a maximum of only a $4\%$ drop in mAP and a maximum of $0.4\%$ gain in mAP compared to their respective teacher models. At the same time, number of trainable parameters is reduced by $85.4\%$ in the case of StudentV1 and $91.16\%$ in the case of StudentV2. A decrease in the number of trainable parameters leads to significant reduction in training time per epoch. We got a nearly $3.1X$ times drop in training time per epoch in the case of StudentV1 and a $1.05X$ times drop in training time per epoch in the case of StudentV2. Once the student-teacher knowledge distillation is done on a dataset, we can reuse our student model for any fine-tuning task on the same dataset without repeating the knowledge distillation step and without training the teacher model on the fine-tuning task. This solution can be used in diverse applications where model compression is required.

%
%
%
%

\end{document}